Proc. (CD) of *IEEE Conference on AI Tools and Engineering* (ICAITE-08), 2008.# Handwritten *Devanagari* Script Segmentation: A Non-linear Fuzzy Approach

Ram Sarkar [1ξ], Bibhash Sen [2], Nibaran Das [3], Subhadip Basu [3]

[1] Computer Sc. & Engg. Dept., MCKV Institute of Engineering, Liluah, Howrah-711204, India.
[2] Cognizant Technology Solutions, Sector V, Kolkata, India.
[3] Computer Sc. & Engg. Dept., Jadavpur University, Kolkata-700032, India.

[ξ] Corresponding Author; e-mail: raamsarkar@gmail.com*Abstract*

*The paper concentrates on improvement of segmentation accuracy by addressing some of the key challenges of handwritten Devanagari word image segmentation technique. In the present work, we have developed a new feature based approach for identification of Matra pixels from a word image, design of a non-linear fuzzy membership functions for headline estimation and finally design of a non-linear fuzzy functions for identifying segmentation points on the Matra. The segmentation accuracy achieved by the current technique is 94.8%. This shows an improvement of performance by 1.8% over the previous technique [1] on a 300-word dataset, used for the current experiment.*## 1. Introduction

Segmentation of documents into lines and words, and words into individual characters and symbols extracted from optically scanned document images of handwritten text, is one of the major problems of optical character recognition (OCR). Extraction and localization of candidate characters, different modified shapes of characters and character components from isolated word images is often significant enough to make a decisive contribution towards the overall performance of the system. The better is the segmentation process, the lesser is the ambiguity encountered in recognition of candidate characters or word pieces.

Word segmentation is one of the core problems of OCR of handwritten text, which has long been an active area of research. Some important contributions so far made in this field include of English texts [3], [4], [5], [6], Chinese script [7], Arabic characters [8], Bengali scripts [2], [9], [10], [12], [13], [14].

In this paper we have considered the problem of segmenting handwritten text in *Devanagari*. The problem of segmenting extracted words into constituent characters is difficult, especially for *Devanagari*, an important East Asian script widely used in India. Many Indian languages including Sanskrit, Marathi and Hindi (the official language of India) use the *Devanagari* script. Several other languages such as Gujarati, Punjabi and Bengali use scripts, which are very similar to *Devanagari*. *Devanagari* is a derivative of ancient of Brahmi, the mother of all Indian scripts. *Devanagari* is more complex than the familiar Roman script in several ways: (a) It has many more basic characters in its alphabet; (b) Vowels are written as medications of the consonants characters.

The work relating to OCR of *Devanagari* script is found to have few references in the literature. Such instances found in [1], [14], [15], [16], [17]. The problem of *Devanagari* text segmentation has been addressed in [1], [14], [17]. In one of our earlier works [1], a fuzzy technique was proposed for segmentation of handwritten *Devanagari* word images. Although the technique of *Devanagari* character segmentation as described in [17], has shown a high success rate by properly segmenting nearly 88% of characters of printed text, it will not be effective for hand written text, where the *Matra*s are not strictly horizontal as those are in printed words.

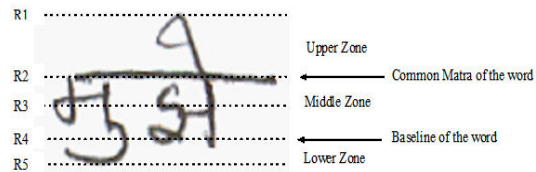

**Fig 1.** Common *Matra* and the three zones and region boundaries of a word image

*Devanagari* character segmentation is predominantly based on detection of an important feature of *Devanagari* text, called the *Matra*. A *Matra* is a horizontal line, which passes touching the upper part of many characters of *Devanagari* script as shown in Fig. 1. Depending on the characters, it covers at most the entire character width. The consecutive characters, in a *Devanagari* word, which have *Matra*s, are joined through a common *Matra* formed by joining the *Matra*s of individual characters. This line may have some discontinuity over the positions where the characters in the word appear without *Matra*s as shown in Fig. 2.

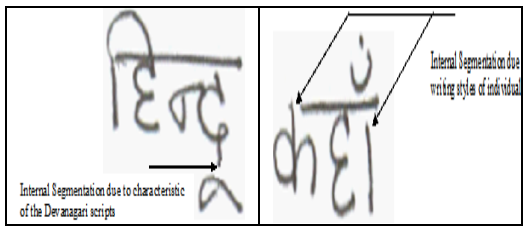

**Fig 2.** Different types of internal segmentation within a handwritten *Devanagari* word image.

Compared to Roman script, some special features of *Devanagari* script make the task of character segmentation complex for words appearing in pieces of *Devanagari* text. There are some characters in *Devanagari* script, called *Modified shapes*, which are not positioned in a strict left to right non-overlapping sequence with adjacent characters in a word. For all of these, segmentation of *Devanagari* words just by observing the valleys in the histogram, drawn by adding the column wise pixel densities of the word image, is not possible for *Devanagari* script.

Appearance of consecutive characters in overlapping column positions over a text line makes the problem of *Devanagari* word segmentation more complex compared to segmentation of English words. The problem becomes compounded with handwritten *Devanagari* words because of variation in sizes and shapes of handwritten characters.

In comparison to our previous work on development of a fuzzy technique for handwritten *Devanagari* word segmentation [1], the current work concentrates on improvement of segmentation accuracy by re-addressing some of the key modules of our previous work. The most significant contributions of the present work are on development of a new feature based approach for identification of *Matra* pixels from a word image, design of non-linear fuzzy membership functions for headline estimation, identification of connected components for further segmentation and finally design of non-linear fuzzy functions for identifying segmentation points on the *Matra*. Following sections briefly describe the key functional modules developed for the current work.

## 2. Noise Elimination

Depending on the data acquisition type, the raw data is subjected to a number of preliminary processing steps to make the data usable. Preprocessing aims to produce data that are easy to operate accurately in document image processing. In the present work, we have used several computing metrics based on spatial attributes of pixels of the binary image. Therefore, noise pixels appearing at the background and along the contour of the word image may affect the segmentation accuracy. To remove a noisy pixel and to smooth the contours of data, we have used a sequence of erosion and dilation, two basic mathematical morphological operators [14], on the input handwritten word images.

## 3. Zone Determination in a Word Image

Word images, written in *Devanagari* script, can be partitioned horizontally into three adjacent zones as shown in Fig. 1. The portion of each word on and above the *Matra* is identified as the 'upper zone'. The main body of the characters in a word and the portion of the word below the main body are identified as the 'middle zone' and the 'lower zone' respectively. So the three adjacent zones of a word image, mentioned before, need to be identified before segmenting it into constituent characters. More specifically, the top row of the upper zone ($R_1$), the top row of the middle zone ($R_2$), the mid line ($R_3$) of the middle zone, the bottom row of the middle zone ($R_4$) and the bottom row of the lower zone ($R_5$) are to be identified first from the word image.

A horizontal pixel scan of the word image from top towards bottom identifies the first row with at least a single black pixel as the top row ($R_1$) of the upper zone. Similarly, another horizontal scan from bottom towards top identifies the first row again with at least a single black pixel as the bottom row ($R_5$) of the lower zone. Identification of the top row ($R_2$) and bottom row ($R_4$) boundaries of the middle zone is a difficult task in handwritten *Devanagari* words.

For each black pixel in each row of the word image, the length of the *longest run* of black pixels in horizontal direction is computed. Then the sum of lengths of the longest runs of all black pixels in each row is computed and plotted, as discussed in [1]. The row with the highest sum represents the row with maximum horizontalness of consecutive black pixels. This row signifies the upper boundary of the middle zone ($R_2$). To identify the lower boundary of the middle zone, the number of transitions from text to background pixels and vice versa is computed along each row, starting from $R_2$. In each row, starting from the bottom of the lower zone to the top of middle zone the sum of all such transitions is computed. The average of these row wise sums, denoted as $n_{ave}$, is then computed, considering all rows from middle and lower zones. Finally, the first row from bottom most line with the value of transition sum exceeding $n_{ave}$, is identified as the lower boundary of the middle zone.

## 4. Headline Estimation: Fuzzy Approach

*Matra* or the common headline of a word image may be identified as the continuous horizontal stripe of black pixels appearing at the top of most of the characters and some of modified shapes in the word. All the component characters and modified shapes appear to touch each other only through the *Matra* of the word. However, in a cursive handwriting the appearance of a *Matra* is often disjoint and wavy. This makes the identification of potential *Matra* pixels a challenging task. In the present work, we have developed two fuzzy measures to identify the membership value of each pixel for its potential of belongingness to *Matra*.

### 4.1. Horizontalness feature

The horizontalness feature of *Devanagari* Script is very unique and appears very prominently in a word images. This horizontalness property of the *Matra* may be extracted from the row wise sum of continuous run of black pixels. This value is normalized with respect to the maximum longest run value of any pixel within the word image.

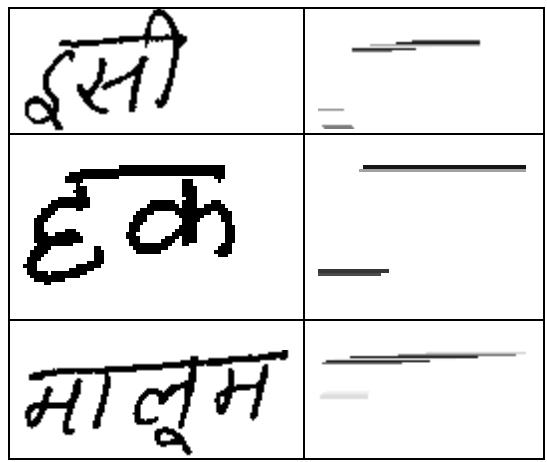

**Fig 3.** Sample Word images and the corresponding horizontal longest run components that exceeds the mean horizontalness of the respective words

### 4.2. Verticalness feature

Many characters and modified shapes in *Devanagari* script have vertical stripe of black pixels, as a part of their shapes. This vertical stripe often appears at the right side, middle or left side of the characters. These stripes touch the *Matra* of a word image and often extend till the bottom of the respective characters or modified shapes. In the present work, we have developed a technique to identify prominent vertical stripes in word image and identify their average top and bottom rows within the principal segments. This verticalness property of the *Matra* may be extracted from the column wise count of continuous run of black pixels. This value is normalized with respect to the maximum longest run value of any pixel within the word image.

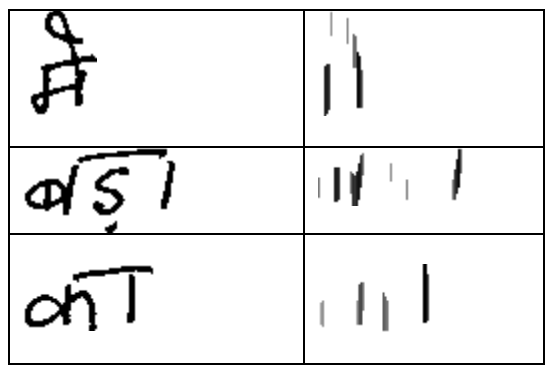

**Fig 4:** Sample Word images and the corresponding vertical longest run components that exceeds the mean verticalness of the respective words

### 4.3. Fuzzy Membership Function for Headline Estimation

We have designed a bell shaped membership functions to map the horizontalness feature values of each row to determine its belongingness in the *Matra* region. The generalized bell function depends on three parameters *a*, *b*, and *c* as given by

$$f(x;a,b,c) = \frac{1}{1+\left|\frac{x-c}{a}\right|^{2b}}$$

where the parameter *b* is usually positive. The parameter *c* locates the center of the curve, i.e., $R_2$ and *x* is the row index for any black pixel $P_{xy}$ in the word image. For computation of the fuzzy feature values, we have designed a fuzzy function, viz, $f_h(x_h, f(x;a,b,c))$ for horizontalness feature respectively. Such that,

$$f_h(x_h, f(x;a,b,c)) = x_h * f(x;a,b,c)$$

Where, $x_h$ is normalized horizontalness component of each pixel $P_{xy}$ under consideration and $0 \leq x_h \leq 1$. Fig. 5 shows a diagrammatic representation of the bell shaped fuzzy membership function, designed for the present work.

A pixel $P_{xy}$ is identified as a headline pixel, if its value exceeds the mean of all such $f_h(P_{xy})$ values within the region $R_1$-$R_3$.

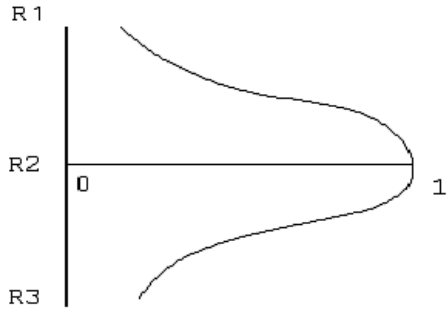

**Fig 5.** Non-linear Fuzzy Bell-shape memberships function for *Matra* determination.

### 5. Fuzzy Segmentation Features

After *Matra* of a word is identified the next task becomes to identify certain column positions on the *Matra* from where the word can be vertically segmented into constituent characters. Such column positions are called terminal points of segments. One of the prominent features for identifying terminal points of segments is the number of black pixels along each vertical column position on the *Matra*. The less is the number of black pixels along a vertical column position on the *Matra*, the higher is its degree of belongingness ($\mu_1$) to the set of terminal segment-points. On this basis a bell-shaped fuzzy membership function ($\mu_1$), as discussed in section 4.3, is designed.

Another feature ($F_2$) is considered here within the region ($R_2 - R_3$). Here again the more is the distance, the less is the degree of belongingness ($\mu_2$) of the associated point to the set of segment terminal points. Detailed description of these three features is already given in [2]. The necessary membership functions ($\mu_1$, $\mu_2$) for these features are shown in Fig. 6.

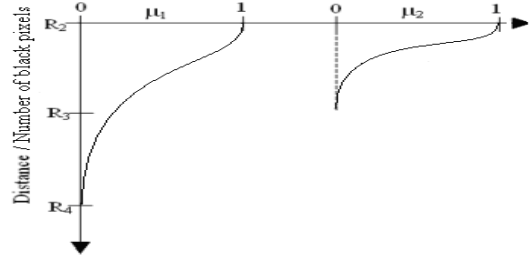

**Fig 6.** A pictorial representation of Fuzzy membership functions $\mu_1$, $\mu_2$

To determine finally whether a black pixel on the *Matra* can be considered as a segment terminal point, the average of all the two feature values exceed certain predetermined threshold, are finally considered as segment terminal points. The threshold is fixed up by taking the average of all the two feature values of all the black pixel positions over the *Matra* of a segmented word.

### 6. Results and discussion

To evaluate the performance of the technique, described here, for segmentation of word images a total of 300 word images are collected from different documents to include varieties in writing styles. The documents are digitized using a flatbed scanner at a resolution of 300 dpi. The digitized documents so prepared are finally binarized simply through thresholding. Due to non-availability of standard datasets for handwritten *Devanagari* word images, the performance of the current technique could not be compared with existing segmentation algorithms described in [11-14]. However, we have compared the performance of the present work with one of our previous works, described in [1].

For designing of the fuzzy function, as shown in section 4.3, the values of two positive constants *a* and *b* were chosen as 2 and 1 respectively. As discussed earlier, the row index of the lower boundary of the upper zone ($R_2$) is assigned to the third constant *c* in the said fuzzy function.

Some of the sample images of *Devanagari* words, which were properly segmented by the present technique, are shown in Fig. 7. Also, some, on which the technique fails at some points, are shown in Fig. 8. In the Figs. 7-8, the pixel positions, which are identified as potential segment points on the *Matra*, are shown with darker shading, where as the other pixel positions on the same *Matra* are shown with lighter shading.

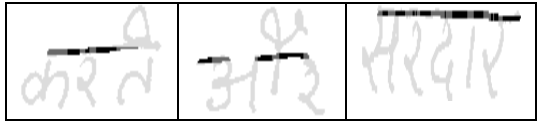

**Fig 7.** Samples of some successfully segmented word Images by Bell-shape fuzzy membership function.

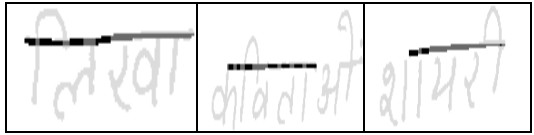

**Fig 8.** Samples of some unsuccessfully segmented word images by Bell-shape fuzzy membership function.

To evaluate the segmentation performance of the present technique the following expression is developed.

$$\text{Success rate} = (C_t / (C_t + C_u)) * 100$$

Where $C_t$ = the number of segment terminal points producing true segmentation and $C_u$ = the number of segment terminal points producing under segmentation. Whether a segment terminal point, identified by the present technique, produces true segmentation or under segmentation is determined through visual observation here. On the basis of this, the success rate of the present technique is computed to be 94.8% out of the 300 word images. On the same word dataset, our previous technique shows a success rate of 93%. This shows an improvement of 1.8% over the technique reported by us in [1]. The improved performance on the same dataset validates our choice of using non-linear fuzzy membership function over the previous choice of triangular membership functions. Also, we observed that the use of horizontalness and vertical ness features refine the selection of headline pixels over the previous technique. Finally, considering the enormous complexity of *Devanagari* script, the contribution of the present approach may be considered significant with satisfactory segmentation performances.

## Acknowledgements

Authors are thankful to the "Center for Microprocessor Application for Training Education and Research", "Project on Storage Retrieval and Understanding of Video for Multimedia" of Computer Science & Engineering Department, Jadavpur University, for providing infrastructure facilities during progress of the work. One of the authors, Mr. Ram Sarkar, is thankful to MCKV Institute of Engineering for kindly permitting him to carry on the research work.